\documentclass[twoside,11pt]{article}

\usepackage{paralist,amsmath, amssymb}
\usepackage{algorithm,algorithmic}
\usepackage{jmlr2e}

\usepackage[colorlinks,
            linkcolor=blue,
            citecolor=blue,
            urlcolor=magenta,
            linktocpage,
            plainpages=false]{hyperref}

\makeatletter
\newcounter{ALC@tempcntr}
\makeatother

\makeatletter
\AtBeginDocument{\Hy@breaklinkstrue}
\makeatother

\newtheorem{thm}{Theorem}

 \newtheorem{ass}{Assumption}

\def \w {\mathbf{w}}
\def \R {\mathbb{R}}

\def \W {\mathcal{W}}
\def \N {\mathcal{N}}

\def \I {\mathcal{I}}

\def \C {\mathcal{C}}

\def \S {\mathcal{S}}
\def \N {\mathbb{N}}

\DeclareMathOperator*{\Reg}{Regret}
\DeclareMathOperator*{\WAReg}{A-Regret}
\DeclareMathOperator*{\SAReg}{SA-Regret}

\begin{document}

\title{Adaptive Regret of Convex and Smooth Functions}

\author{\name Lijun Zhang \email zhanglj@lamda.nju.edu.cn\\
       \addr National Key Laboratory for Novel Software Technology\\
       Nanjing University, Nanjing 210023, China\\
       \name Tie-Yan Liu \email tie-yan.liu@microsoft.com\\
       \addr Microsoft Research Asia, Beijing 100080, China  \\
\name Zhi-Hua Zhou \email zhouzh@lamda.nju.edu.cn\\
       \addr National Key Laboratory for Novel Software Technology\\
       Nanjing University, Nanjing 210023, China
}
\editor{}

\maketitle

\begin{abstract}
We investigate online convex optimization in changing environments, and choose the adaptive regret as the performance measure. The goal is to achieve a small regret over \emph{every} interval so that the comparator is allowed to change over time. Different from previous works that only utilize the convexity condition, this paper further exploits smoothness to improve the adaptive regret. To this end, we develop novel adaptive algorithms for convex and smooth functions, and establish problem-dependent regret bounds over any interval. Our regret bounds are comparable to existing results in the worst case, and become much tighter when the comparator has a small loss.
\end{abstract}


\section{Introduction}
Online convex optimization (OCO) is a powerful learning framework which has both theoretical and practical appeals \citep{zinkevich-2003-online}. Given a convex decision set $\W$, the learner is required to select a decision $\w_t \in \W$ in each round $t$. Then, a convex loss function $f_t:\W \mapsto \R$ is revealed, and the learner suffers loss $f_t(\w_t)$. The goal is to minimize the cumulative loss of the online learner, or equivalently the regret defined as
\[
\Reg=\sum_{t=1}^T f_t(\w_t) - \min_{\w \in \W} \sum_{t=1}^T f_t(\w)
\]
which is the difference of losses between the learner and the optimal solution in hindsight. In the past decades, various algorithms for minimizing the regret have been developed \citep{Online:suvery,Intro:Online:Convex}.

OCO is a natural choice for changing environments in the sense that the loss arrives dynamically. However, in the real-world application, we are also facing another dynamic challenge---the optimal solution may change continuously.  For example, in online recommendation, $\w$ models the interest of users, which could evolve over time. In this scenario, regret is no longer a suitable measure of performance, since the online learner is compared against a \emph{fixed} decision. So, the traditional regret is also  referred to as \emph{static} regret to emphasize that the comparator is static.

To cope with changing environments,  the notion of adaptive regret has been proposed and received considerable interests  \citep{Adaptive:Hazan,Adaptive:ICML:15,Dynamic:Regret:Adaptive}. The key idea is to minimize the ``local'' regret
\[
\Reg\big([r, s]\big)=\sum_{t=r}^{s} f_t(\w_t) - \min_{\w \in \W} \sum_{t=r}^{s} f_t(\w)
\]
of \emph{every} interval $[r, s] \subseteq [T]$. Requiring a low regret over any interval essentially means the online learner is evaluated against a changing comparator. For convex functions, the state-of-the-art algorithm achieves an $O(\sqrt{(s-r) \log s})$  regret over any interval $[r, s]$ \citep{Improved:Strongly:Adaptive}, which is close to the minimax regret over a fixed interval \citep{Minimax:Online}. In the studies of static regret, it is well-known that the regret bound can be improved when additional curvatures, such as smoothness, are present \citep{NIPS2010_Smooth}. Thus, it is natural to ask whether smoothness can also be exploited to enhance the adaptive regret. This paper provides an affirmative answer by developing adaptive algorithms for convex and smooth functions that enjoy tighter bounds.

We remark that directly combining the regret of convex and smooth functions with existing adaptive algorithms does not give a tight adaptive regret, because of the following technical challenges.
\begin{compactitem}
  \item The regret bound for convex and smooth functions requires to know the loss of the optimal decision \citep{NIPS2010_Smooth}, which is generally unavailable.
  \item Existing adaptive algorithms have some components, including a meta-algorithm and a set of intervals, that cannot utilize smoothness.
\end{compactitem}

To address the above challenges, we first introduce the scale-free online gradient descent (SOGD), a special case of the scale-free mirror descent \citep{Scale:Free:Online}, and demonstrate that SOGD is able to exploit smoothness automatically  and does not need any prior knowledge. Then, we develop a Strongly Adaptive algorithm for Convex and Smooth functions (SACS), which runs multiple instances of SOGD over a set of carefully designed intervals, and combines them with an expert-tracking algorithm that can benefit from small losses.  Let $L_r^s=\min_{\w \in \W} \sum_{t=r}^{s} f_t(\w)$ be the minimal loss over an interval $[r,s]$. Our theoretical analysis demonstrates that the regret of SACS over any interval $[r,s]$ is $O(\sqrt{L_r^s \log s  \cdot \log (s-r)})$, which could be much smaller than the existing $O(\sqrt{(s-r) \log s})$ bound when $L_r^s$ is small. Finally, to further improve the performance, we propose a novel way to construct problem-dependent intervals, and attain an $O(\sqrt{L_r^s \log L_1^s  \cdot \log L_r^s})$ bound.
\section{Related Work}
Adaptive regret has been studied in the settings of prediction with expert advice (PEA) and online convex optimization (OCO). Existing algorithms are closely related in the sense that adaptive algorithms designed for OCO are usually built upon those designed for PEA.

In an early study of PEA, \citet{LITTLESTONE1994212} develop one variant of weighted majority algorithm for tracking the best expert. One intermediate result, i.e.,~Lemma 3.1 of \citet{LITTLESTONE1994212} provides a mistake bound for any interval, which is analogous to the adaptive regret. The concept of adaptive regret is formally introduced by \citet{Adaptive:Hazan} in the context of OCO. Specifically, \citet{Adaptive:Hazan} introduce the adaptive regret
\begin{equation}\label{eqn:weak:regret}
\WAReg(T) =\max_{[r, s] \subseteq [T]} \Reg\big([r, s]\big)
\end{equation}
which is the maximum regret over any contiguous interval, and propose a new algorithm named follow the leading history (FLH), which contains $3$ parts:
\begin{compactitem}
  \item An expert-algorithm, which is able to minimize the static regret of a given interval;
  \item A set of intervals, each of which is associated with an expert-algorithm that minimizes the regret of that interval;
  \item A meta-algorithm, which combines the predictions of active experts in each round.
\end{compactitem}

For exponentially concave (abbr.~exp-concave) functions, \citet{Adaptive:Hazan} use online Newton step \citep{ML:Hazan:2007} as the expert-algorithm. For the construction of intervals, they consider two different approaches. In the first approach, the set of intervals is $\{ [t, \infty], t \in \N \}$ which means an expert will be initialized at each round $t$ and live forever. In the second approach, the set of intervals is $\{ [t, e_t], t \in \N \}$, meaning the expert that becomes active in round $t$ will be removed after $e_t$. Here, $e_t$ denotes the ending time of the interval started from $t$, and its value is set according to a data streaming algorithm. \citet{ML:Hazan:2007} develop a meta-algorithm based on Fixed-Share \citep{Herbster1998}, and allow the set of experts to change dynamically.

FLH with the first set of intervals attains an $O(d \log T)$ adaptive regret, where $d$ is the dimensionality, but is inefficient since it maintains $t$ experts in round $t$. In contrast, FLH with the second set of intervals achieves a higher $O(d \log^2 T)$ bound, but is efficient because it only keeps $O(\log t)$ experts in the $t$-th round. Thus, we observe that the intervals control the tradeoff between the adaptive regret and the computational cost.  On one hand, the interval set should be large so that for every possible interval there exists an expert that works well. On the other hand, the number of intervals should be small, since running many experts in parallel will result in high computation cost.
\citet{Track_Large_Expert} and \citet{Dynamic:Regret:Adaptive} have developed new ways to construct intervals which  can trade effectiveness for efficiency explicitly. Furthermore, when the function is strongly convex, the dependence on $d$ in the upper bound disappears \citep{Dynamic:Regret:Adaptive}.

For convex functions, \citet{ML:Hazan:2007}  modify the FLH algorithm by replacing the expert-algorithm with any low-regret method for convex functions, and introducing a parameter of step size in the meta-algorithm. In this case, the efficient and inefficient versions of FLH achieve $O(\sqrt{T \log^3 T})$ and $O(\sqrt{T \log T})$  adaptive regret bounds, respectively.\footnote{As pointed out by \citet{Hazan:2009:ELA}, online gradient descent with constant step size \citep{zinkevich-2003-online} can also be used to minimize the adaptive regret of convex functions, and the bound is $O(\sqrt{T})$.} One limitation of this result is that it does not guarantee to perform well on small intervals, because the upper bounds are meaningless for intervals of size $O(\sqrt{T})$.

The adaptive regret of PEA setting is studied by \citet{Adamskiy2012}. Let $L_{t,i}$ be the loss of the $i$-th expert in round $t$, and $L_t$ be the loss of the learner, which is generally a convex combination of $L_{t,i}$'s. In this case, the regret over interval $[r, s]$ in (\ref{eqn:weak:regret}) becomes
\[
\Reg\big([r, s]\big)=\sum_{t=r}^s L_t - \min_{i} \sum_{t=r}^s L_{t,i}.
\]
They pointed out that the meta-algorithm of \citet{ML:Hazan:2007} can be reduced to the Fixed-Share algorithm with a special configuration of parameters. Although  Fixed-Share is designed to minimize the tracking regret, \citet{Adamskiy2012} show that it can also minimize the adaptive regret. Combining Hoeffding bound \citep[Lemma 2.2]{bianchi-2006-prediction} and (1a) of \citet{Adamskiy2012},  it is easy to prove that the adaptive regret of Fixed-Share is $O(\sqrt{T \log NT})$, where $N$ is the number of experts.\footnote{We need to use Hoeffding bound to convert the mix loss defined by \citet{Adamskiy2012} to the traditional weighted loss.} Unfortunately, it also does not respect short intervals well.

\begin{figure*}
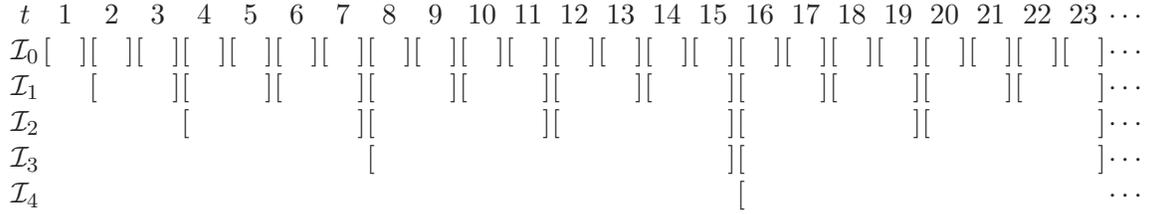

\centering
\begin{tabular}{@{}c@{\hspace{0.2ex}}*{23}{@{\hspace{0.1ex}}c}@{\hspace{0.2ex}}c@{}}
$t$ & 1 & 2 & 3 & 4 & 5 & 6 & 7 & 8 & 9 & 10 &11 &12 & 13 & 14 &15 & 16 & 17 & 18&19 &20 &21 &22 & 23 &$\cdots$ \\
 $\I_0$ & [\quad ] & [\quad ] &  [\quad  ] & [\quad ] & [\quad ] & [\quad ] & [\quad ] & [\quad ] & [\quad ] & [\quad ] & [\quad ] &[\quad ] &[\quad ] & [\quad ] & [\quad ] & [\quad ]& [\quad ]&[\quad ] & [\quad ]& [\quad ]& [\quad ]& [\quad ]& [\quad ]& $\cdots$   \\
 $\I_1$ &  & [\quad  \phantom{]}& \phantom{[}\quad ] & [\quad \phantom{]} & \phantom{[}\quad ] & [\quad \phantom{]} & \phantom{[}\quad ] & [\quad \phantom{]} & \phantom{[}\quad ] & [\quad \phantom{]} & \phantom{[}\quad ] &[\quad \phantom{]} &\phantom{[}\quad ] & [\quad \phantom{]} & \phantom{[}\quad ] & [\quad \phantom{]} & \phantom{[}\quad ]& [\quad \phantom{]} & \phantom{[}\quad ]& [\quad \phantom{]}& \phantom{[}\quad ]&[\quad \phantom{]} &\phantom{[}\quad ]& $\cdots$   \\
 $\I_2$  &  & &  & [\quad \phantom{]} & \phantom{[}\quad \phantom{]} & \phantom{[}\quad \phantom{]} & \phantom{[}\quad ] & [\quad \phantom{]} & \phantom{[}\quad \phantom{]} & \phantom{[}\quad \phantom{]} & \phantom{[}\quad ] &[\quad \phantom{]} &\phantom{[}\quad \phantom{]} & \phantom{[}\quad \phantom{]} & \phantom{[}\quad ] & [\quad \phantom{]} & \phantom{[}\quad \phantom{]}&\phantom{[}\quad \phantom{]} &\phantom{[}\quad ] &[\quad \phantom{]} &\phantom{[}\quad \phantom{]} &\phantom{[}\quad \phantom{]} & \phantom{[}\quad ] &$\cdots$   \\
  $\I_3$   &  & &  &  &  &  &  & [\quad \phantom{]} & \phantom{[}\quad \phantom{]} & \phantom{[}\quad \phantom{]} & \phantom{[}\quad \phantom{]} &\phantom{[}\quad \phantom{]} &\phantom{[}\quad \phantom{]} & \phantom{[}\quad \phantom{]} & \phantom{[}\quad ] & [\quad \phantom{]} & \phantom{[}\quad \phantom{]}& & & & & & \phantom{[}\quad ]&$\cdots$   \\
$\I_4$   &  & &  &  &  &  &  &  &  &  &  & & &  &  & [\quad \phantom{]} & & & & & & & &$\cdots$
\end{tabular}\vspace{-2ex}
\caption{Geometric covering (GC) intervals of \citet{Adaptive:ICML:15}. In the figure, each interval is denoted by $[ \quad ]$.}
\label{fig:interval:saol}
\end{figure*}
To ensure a good performance on every interval, \citet{Adaptive:ICML:15} propose the notion of strongly adaptive regret
\[
\SAReg(T,\tau) = \max_{[s, s+\tau -1] \subseteq [T]} \Reg\big([s, s+\tau -1]\big)
\]
which emphasizes the dependency on the interval length $\tau$, and investigate both the PEA and OCO settings. The main contribution of that paper is a new meta-algorithm for combining experts, namely strongly adaptive online learner (SAOL), which is similar to the multiplicative weights method \citep{v008a006}. Furthermore, they also propose a different way to construct the set of intervals as
\[
 \I= \bigcup_{k \in \N \cup \{0\}} \I_k
 \]
 where for all $k \in \N \cup \{0\}$
 \[
 \I_k=\left\{ [ i \cdot 2^k, (i+1) \cdot 2^k -1]: i \in \N\right\}.
 \]
 Following \citet{Improved:Strongly:Adaptive}, we refer to $\I$ as geometric covering (GC) intervals and present a graphical illustration in Fig.~\ref{fig:interval:saol}. It is obvious to see that each $\I_k$ is a partition of $\N \setminus \{1,\cdots, 2^k-1 \}$ to consecutive intervals of length $2^k$.

In the PEA setting, by using multiplicative weights as the expert-algorithm, \citet{Adaptive:ICML:15} establish a strongly adaptive regret of $O(\sqrt{\tau \log N } + \log T \sqrt{\tau})$. In the OCO setting,  by using online gradient descent as the expert-algorithm,  \citet{Adaptive:ICML:15} establish a strongly adaptive regret of $O(\log T \sqrt{\tau})$. Those rates are further improved by  \citet{Improved:Strongly:Adaptive}, who develop a new meta-algorithm named as sleeping coin betting (CB). The strongly adaptive regrets  of PEA and OCO are improved to $O(\sqrt{\tau \log N T })$ and $O( \sqrt{\tau \log T} )$, respectively. Recently, \citet{Adaptive:One:Gradient} demonstrate that for minimizing the adaptive regret of convex functions, we can use surrogate loss to reduce the number of gradient evaluations per round from $O(\log T)$ to $1$.

After personal discussions in ICML 2019, we realize that \citet{jun2017} have analyzed the adaptive regret of convex and smooth functions in the journal version of \citet{Improved:Strongly:Adaptive}. In particular, they use the AdaptiveNormal potential \citep{NIPS2017_6811} in sleeping CB, and demonstrate that the regret of convex and smooth functions over any interval $[r,s]$ can be upper bounded by $O(\log s  \sqrt{L_r^s  })$ \citep[Corollary 10]{jun2017}. Although our first result, i.e., the $O(\sqrt{L_r^s \log s \cdot \log (s-r)})$ regret over $[r,s]$, is similar to theirs, the corresponding algorithm is different. Furthermore, our problem-dependent intervals and the second result, i.e., the $O(\sqrt{L_r^s \log L_1^s  \cdot \log L_r^s})$ regret over $[r,s]$, are novel.

Finally, we note that adaptive regret is closely related to the tracking regret in PEA \citep{Herbster1998,Track_Large_Expert,Fixed:Share:NIPS12} and dynamic regret in OCO \citep{Dynamic:ICML:13,Dynamic:AISTATS:15,Dynamic:Strongly,Dynamic:2016,Dynamic:Regret:Squared,Adaptive:Dynamic:Regret:NIPS}. Specifically, from adaptive regret, we can derive a tight bound for the tracking regret \citep{Improved:Strongly:Adaptive} and a special form of dynamic regret \citep{Dynamic:Regret:Adaptive}.
\section{Main Results}
We first investigate how to utilize smoothness to improve the static regret, then develop a strongly adaptive algorithm for convex and smooth functions, and finally propose data-dependent intervals to further strengthen the performances.

\subsection{Scale-free Online Gradient Descent (SOGD)}
We introduce common assumptions used in our paper.

\begin{ass}\label{ass:3} The domain $\W$ is convex, and its diameter is bounded by $D$, i.e.,
\begin{equation}\label{eqn:domain}
\max_{\w, \w' \in \W} \|\w -\w'\|_2 \leq D.
\end{equation}
\end{ass}

\begin{ass}\label{ass:4} All the online functions are convex and nonnegative.
\end{ass}

\begin{ass}\label{ass:5} All the online functions are $H$-smooth over $\W$, that is,
\begin{equation} \label{eqn:f:smooth}
\left \|\nabla f_t(\w)-\nabla f_t(\w') \right\| \leq H \|\w-\w'\|
\end{equation}
for all $\w, \w' \in \W$, $t \in [T]$.
\end{ass}

Note that in Assumption~\ref{ass:4}, we require the online function to be nonnegative outside the domain $\W$. This is a precondition for establishing the self-bounding property of smooth functions, which can be exploited to deliver a tight regret bound. Specifically,  \citet{NIPS2010_Smooth} consider online gradient descent with constant step size:
\[
\w_{t+1} = \Pi_{\W}\big[\w_t - \eta \nabla f_t(\w_t)\big], \ \forall t \geq 1
\]
where $\w_1 \in \W$ and  $\Pi_{\W}[\cdot]$ denotes the projection onto the nearest point in $\W$, and prove the following regret bound \citep[Theorem 2]{NIPS2010_Smooth}.
\begin{thm} \label{thm:smooth:1} Let $B \geq 0$ and $L \geq 0$ be two constants,  set the step size in OGD as
\[
\eta = \frac{1}{H B^2 + \sqrt{H^2B^4+ H B^2 L}},
\]
and $\w_1 =0$. Under Assumptions~\ref{ass:4} and \ref{ass:5}, we have
\[
\sum_{t=1}^T f_t(\w_{t}) -  \sum_{t=1}^T  f_t(\w)  \leq 4 H B^2 + 2 \sqrt{H B^2 L}
\]
for any $\w \in \W$ such that $\frac{\|\w\|^2}{2} \leq B^2$, and  $\sum_{t=1}^T  f_t(\w) \leq L$.
\end{thm}
The above theorem indicates that under the smoothness condition, the regret bound could be tighter if the cumulative loss of the comparator $\w$ is small, Specifically, when $L=o(T)$, the regret bound becomes $o(\sqrt{T})$, thus improves the minimax rate of online convex optimization \citep{Minimax:Online}. However, one limitation of Theorem~\ref{thm:smooth:1} is that the step size depends on the bound $L$ on the loss in hindsight.

\begin{algorithm}[tb]
   \caption{Scale-free online gradient descent (SOGD)}
   \label{alg:SOGD}
\begin{algorithmic}[1]
   \STATE {\bfseries Input:} parameters $\delta$ and $\alpha$
   \STATE Initialize $\w_1 \in \W$ arbitrarily
   \FOR{$t=1$ {\bfseries to} $T$}
   \STATE Submit $\w_t$ and then receive function $f_t(\cdot)$
   \STATE Suffer loss $f_t(\w_t)$ and set $\eta_t$ as (\ref{eqn:eta_t})
   \STATE Update the decision according to
\[
\w_{t+1} = \Pi_{\W}\big[\w_t - \eta_t \nabla f_t(\w_t)\big]
\]
   \ENDFOR
\end{algorithmic}
\end{algorithm}
The standard way to address the above problem is the ``doubling trick'' \citep{bianchi-2006-prediction}, but it requires the online learner to evaluate the minimal cumulative loss on the fly, which is computationally expensive. Instead, we make use of the scale-free mirror descent algorithm of  \citet{Scale:Free:Online} and set the step size of the $t$-th iteration  as
\begin{equation} \label{eqn:eta_t}
   \eta_t= \frac{\alpha}{\sqrt{\delta+ \sum_{i=1}^t \|\nabla f_i(\w_i)\|^2} }
\end{equation}
where the parameter $\delta>0$ is introduced to avoid being divided by $0$, and $\alpha>0$ is used to fine-tune the upper bound. We note that the step size in (\ref{eqn:eta_t}) is similar to the self-confident tuning originally proposed  for online linear regression  \citep{AUER200248}, and later extended to self-bounded functions  \citep[Theorem 2]{Shai:thesis}. The new algorithm is named as  scale-free online gradient descent (SOGD), and summarized in Algorithm~\ref{alg:SOGD}.

Next, we prove the regret bound of SOGD in the following theorem, which demonstrates that SOGD can make use of smoothness automatically.
\begin{thm} \label{thm:smooth:2} Set $\delta>0$ and $\alpha= D/\sqrt{2}$ in Algorithm~\ref{alg:SOGD}. Under  Assumptions~\ref{ass:3}, \ref{ass:4} and \ref{ass:5}, SOGD satisfies
\[
\begin{split}
\sum_{t=1}^T f_t(\w_{t}) -  \sum_{t=1}^T  f_t(\w)  \leq    8 H D^2 +D\sqrt{2 \delta + 8 H \sum_{t=1}^T  f_t(\w)}
\end{split}
\]
for any $\w \in \W$.
\end{thm}
\textbf{Remark:} First, comparing Theorem~\ref{thm:smooth:2} with Theorem~\ref{thm:smooth:1}, we observe that the regret bound of SOGD is of the same order as that of SGD with optimal parameters. Second, because the step size of SOGD is automatically tuned during the learning process, it is equipped with an anytime regret bound, i.e., its regret bound holds for any $T$. This nice property of SOGD will be utilized to simplify the design of adaptive algorithms.

\subsection{A Strongly Adaptive Algorithm}
Similar to previous studies \citep{Adaptive:Hazan,Adaptive:ICML:15,Improved:Strongly:Adaptive}, our strongly adaptive algorithm contains $3$ components: an expert-algorithm, a set of intervals, and a meta-algorithm.
\begin{figure*}
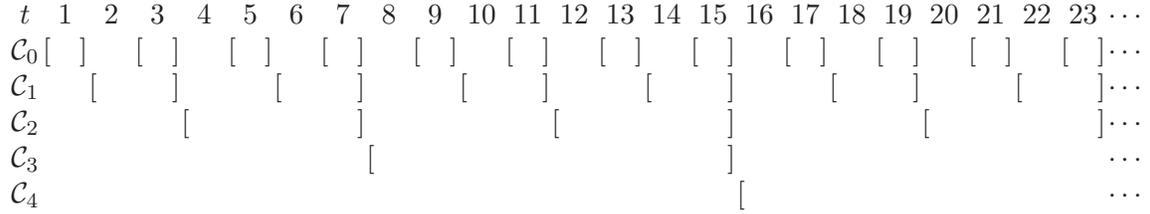

\centering
\begin{tabular}{@{}c@{\hspace{0.2ex}}*{23}{@{\hspace{0.1ex}}c}@{\hspace{0.2ex}}c@{}}
$t$ & 1 & 2 & 3 & 4 & 5 & 6 & 7 & 8 & 9 & 10 &11 &12 & 13 & 14 &15 & 16 & 17 & 18&19 &20 &21 &22 & 23 &$\cdots$ \\
 $\C_0$ & [\quad ] &  &  [\quad ] &  & [\quad ] &  & [\quad ] & & [\quad ] &  & [\quad ] & &[\quad ] &  & [\quad ] & & [\quad ]&& [\quad ]& & [\quad ]& & [\quad ]& $\cdots$   \\
 $\C_1$ &  & [\quad  \phantom{]}& \phantom{[}\quad ] &  &  & [\quad \phantom{]} & \phantom{[}\quad ] &  & & [\quad \phantom{]} & \phantom{[}\quad ] & && [\quad \phantom{]} & \phantom{[}\quad ] &  & & [\quad \phantom{]} & \phantom{[}\quad ]& & &[\quad \phantom{]} &\phantom{[}\quad ]& $\cdots$   \\
 $\C_2$  &  & &  & [\quad \phantom{]} & \phantom{[}\quad \phantom{]} & \phantom{[}\quad \phantom{]} & \phantom{[}\quad ] &  & & & &[\quad \phantom{]} &\phantom{[}\quad \phantom{]} & \phantom{[}\quad \phantom{]} & \phantom{[}\quad ] & & & & &[\quad \phantom{]} &\phantom{[}\quad \phantom{]} &\phantom{[}\quad \phantom{]} & \phantom{[}\quad ] &$\cdots$   \\
  $\C_3$   &  & &  &  &  &  &  & [\quad \phantom{]} & \phantom{[}\quad \phantom{]} & \phantom{[}\quad \phantom{]} & \phantom{[}\quad \phantom{]} &\phantom{[}\quad \phantom{]} &\phantom{[}\quad \phantom{]} & \phantom{[}\quad \phantom{]} & \phantom{[}\quad ] &  && & & & & & &$\cdots$   \\
$\C_4$   &  & &  &  &  &  &  &  &  &  &  & & &  &  & [\quad \phantom{]} & & & & & & & &$\cdots$
\end{tabular}\vspace{-2ex}
\caption{Compact geometric covering (CGC) intervals. In the figure, each interval is denoted by $[ \quad ]$.}
\label{fig:interval:compact}
\end{figure*}
\subsubsection{The Procedure}
For the expert-algorithm, we choose the scale-free online gradient descent (SOGD) in Algorithm~\ref{alg:SOGD}, since it can utilize smoothness to improve the regret bound. For the set of intervals, we can directly re-use the GC intervals of \citet{Adaptive:ICML:15}. However, because an instance of SOGD will be created for each interval and SOGD has an anytime regret bound, we can further simplify GC intervals based on the following observation:   For intervals with the same starting point, we only need to keep the longest one, since the expert associated with this interval can replace others.

Take the set of intervals $\{[4,4], [4, 5], [4, 7]\}$ in Fig.~\ref{fig:interval:saol} as an example, and denote the expert associated with interval $I$ as $E_I$. The expert $E_{[4,7]}$ performs exactly the same as the expert $E_{[4,4]}$ in round $4$, and exactly the same as the expert  $E_{[4,5]}$ in rounds $4$ and $5$. Thus, we can use $E_{[4,7]}$ to replace $E_{[4,4]}$ and  $E_{[4,5]}$ in any place (algorithm or analysis) they appear. Mathematically, our compact geometric covering (CGC) intervals are defined as
\begin{equation}\label{eqn:CGC}
 \C= \bigcup_{k \in \N \cup \{0\}} \C_k
 \end{equation}
 where for all $k \in \N \cup \{0\}$
 \[
 \C_k=\left\{ [ i \cdot 2^k, (i+1) \cdot 2^k -1]: i \textrm{ is odd}\right\}.
 \]
A graphical illustration of CGC intervals is given in Fig.~\ref{fig:interval:compact}. Comparing Fig.~\ref{fig:interval:saol} with Fig.~\ref{fig:interval:compact}, the main difference is that CGC only adds $1$ interval in each round, while CG may add multiple intervals in each round.

Finally, we need to specify the meta-algorithm. One may attempt to use the SAOL  of \citet{Adaptive:ICML:15} or the sleeping CB of \citet{Improved:Strongly:Adaptive}. However, neither of them meets our requirements, because their meta-regret depends on the length of the interval and  cannot benefit from small losses of experts. To avoid this limitation, we can use the AdaNormalHedge \citep{pmlr-v40-Luo15}  or the sleeping CB with the AdaptiveNormal potential \citep{jun2017}, because 
\begin{compactenum}[(i)]
  \item they achieve a small regret when the comparator has a small loss, and thus can be naturally combined with SOGD which enjoys a similar property;
  \item they support the sleeping expert problem, and thus the number of experts can vary over time.
\end{compactenum}
In the following, we choose the AdaNormalHedge \citep{pmlr-v40-Luo15} due to its simplicity. The key ingredients of AdaNormalHedge are a potential function:
\[
\Phi(R,C)=\exp\left(\frac{[R]_+^2}{3C}\right)
\]
where $[x]_+=\max(0,x)$ and $\Phi(0,0)$ is defined to be $1$, and a weight function with respect to this potential:
\[
w(R,C)= \frac{1}{2} \big(\Phi(R+1,C+1) - \Phi(R-1,C+1)\big).
\]
In the $t$-th round, AdaNormalHedge assigns a weight $p_{t,i}$ to an expert $E_i$ according to
\[
p_{t,i} \propto  w(R_{t-1,i},C_{t-1,i})
\]
where $R_{t-1,i}$ is the regret with respect to $E_i$ over the first $t-1$ iterations, and $C_{t-1,i}$ is the sum of the absolute value of the instantaneous regret over the first $t-1$ iterations.

Putting everything together, we present our Strongly Adaptive algorithm for Convex and Smooth functions (SACS) in Algorithm~\ref{alg:SACS}. For each interval $[i,j] \in \C$, we will create an expert $E_{[i,j]}$ which is active during the interval $[i,j]$. Note that in our CGC intervals, the starting point of each interval is \emph{unique}. So, to simplify  notations, we use $E_i$ as a shorthand of $E_{[i,j]}$.

On the $t$-round, we first create an expert $E_t$ by running an instance of SOGD (Step 2) and add it to the set of active experts, denoted by $\S_t$ (Step 3). In Step 4, we receive the prediction $\w_{t,i}$ of each $E_i \in \S_t$, and assign the following weight to $E_i$
\begin{equation} \label{eqn:weight}
   p_{t,i}= \frac{w(R_{t-1,i},C_{t-1,i})}{\sum_{E_i \in \S_t} w(R_{t-1,i},C_{t-1,i}) }
\end{equation}
where
\[
R_{t-1,i} = \sum_{u=i}^{t-1} f_u (\w_{u}) - f_u (\w_{u,i}), \textrm{ and }C_{t-1,i} = \sum_{u=i}^{t-1} \left|f_u (\w_{u}) - f_u (\w_{u,i})\right|.
\]
In Step 5, SACS submits the weighted average of $\w_{t,i}$
\begin{equation} \label{eqn:wt}
  \w_t = \sum_{E_i \in \S_t} p_{t,i}  \w_{t,i}
\end{equation}
as the output, and suffers loss $f_t(\w_t)$. In Step 6, we remove all the experts whose ending times are $t$, and in Step 7, we update the parameters of each remaining expert. Finally, we pass the loss function $f_t(\cdot)$ to all experts in $\S_t$ so that they can update their predictions for the $(t+1)$-th round (Step 8).

\begin{algorithm}[tb]
   \caption{Strongly Adaptive algorithm for Convex and Smooth functions (SACS)}
   \label{alg:SACS}
\begin{algorithmic}[1]
   \FOR{$t=1$ {\bfseries to} $T$}
   \STATE Initialize an expert $E_t$ by invoking SOGD in Algorithm~\ref{alg:SOGD}  and set $R_{t-1,t} =  C_{t-1,t}=0$
   \STATE Add $E_t$ to the set of active experts
   \[
      \S_t=\S_{t-1} \cup \{E_t \}
   \]
   \STATE Receive the prediction $\w_{t,i}$ of each expert $E_i \in \S_t$, and calculate its weight $p_{t,i}$ according to (\ref{eqn:weight})
 \STATE Submit $\w_t$ defined in (\ref{eqn:wt}) and then receive  $f_t(\cdot)$
   \STATE Remove experts whose ending times are $t$
    \[
    \S_{t} = \S_{t} \setminus \{ E_i | [i,t] \in \C \}
    \]
   \STATE For each $E_i \in \S_t$, update
   \[
   R_{t,i} = R_{t-1,i} + f_t(\w_t) - f_t(\w_{t,i}), \textrm{ and }     C_{t,i} = C_{t-1,i} + \left|f_t(\w_t) - f_t(\w_{t,i})\right|
   \]
   \STATE Pass $f_t(\cdot)$ to each expert $E_i \in \S_t$
   \ENDFOR
\end{algorithmic}
\end{algorithm}
\subsubsection{Theoretical Guarantees}
In the following, we present theoretical guarantees of SACS. To simplify our presentations, we assume all the convex functions are bounded by $1$.
\begin{ass}\label{ass:6} The value of each function belongs to $[0,1]$, i.e.,
\[
0 \leq f_t(\w) \leq 1,  \ \forall \w\in \W, t \in [T].
\]
\end{ass}
As long as the loss functions are bounded, they can always be scaled and restricted to $[0, 1]$.

We start with the meta-regret of SACS with respect to an expert $E_i$.
\begin{lemma}\label{lem:meta} Under Assumptions \ref{ass:4} and \ref{ass:6}, for any interval $[i, j] \in \C$, and any $t \in [i,j]$, SACS satisfies
\[
\sum_{u=i}^{t}  \left[f_u(\w_u)-f_u(\w_{u,i})\right] \leq c(t) + \sqrt{2 c(t)\sum_{u=i}^{t} f_u(\w_{u,i})}
\]
where $c(t)=3 \ln(4t^2)$.
\end{lemma}
\textbf{Remark:} First, compared with the meta-regret of SAOL \citep{Adaptive:ICML:15} and sleeping CB  \citep{Improved:Strongly:Adaptive}, the main advantage of SACS is that its upper bound depends on the cumulative loss of the expert, which could be much tighter when the problem is easy. Second, the theoretical guarantee of SACS is an anytime regret bound, since the upper bound holds for any $t \in [i,j]$. Finally, we note that a similar guarantee can be achieved by using the AdaptiveNormal potential in sleeping CB \citep[Lemma 7]{jun2017}. 

Combining Lemma~\ref{lem:meta} with the regret bound of SOGD in Theorem~\ref{thm:smooth:2}, we immediately obtain the following regret bound of SACS over any interval $[i,j] \in \C$.
\begin{lemma}\label{lem:regret:special} Under Assumptions~\ref{ass:3}, \ref{ass:4}, \ref{ass:5} and \ref{ass:6}, for any interval $[i, j] \in \C$,  any $t \in [i,j]$, and any $\w \in \W$, SACS satisfies
\[
\sum_{u=i}^{t}  \left[ f_u(\w_u)-   f_u(\w) \right] \leq a(t)   +\sqrt{b(t) \sum_{u=i}^t    f_u(\w)}
\]
where
\begin{equation} \label{eqn:at}
a(t)= \frac{9}{2} \ln(4t^2) + 18 H D^2  + 2D\sqrt{2 \delta}
\end{equation}
and
\begin{equation} \label{eqn:bt}
b(t)= 24\ln(4t^2)+16 H D^2.
\end{equation}
\end{lemma}

By utilizing the special structure of the interval set $\C$, we extend Lemma~\ref{lem:regret:special} to any interval $[r,s] \subseteq [T]$.
\begin{thm} \label{thm:adap:regret} Under Assumptions~\ref{ass:3}, \ref{ass:4}, \ref{ass:5} and \ref{ass:6}, for  any interval $[r,s] \subseteq [T]$ and any $\w \in \W$, SACS satisfies
\[
\begin{split}
\sum_{t=r}^{s} \left[ f_t(\w_t) -  f_t(\w) \right] \leq  v a(s) + \sqrt{v b(s) \sum_{t=r}^{s}  f_t(\w)} =  O\left(  \sqrt{\left(\sum_{t=r}^{s}  f_t(\w) \right)\log s  \cdot \log (s-r)}  \right)
\end{split}
\]
where $v \leq \lceil \log_2 (s-r+2)\rceil$, $a(\cdot)$ and $b(\cdot)$ are respectively defined in (\ref{eqn:at}) and (\ref{eqn:bt}).
\end{thm}
\textbf{Remark:} In the literature, the best adaptive regret  for convex functions is $O( \sqrt{(s-r) \log s})$ of \citet{Improved:Strongly:Adaptive}. Although our upper bound in Theorem~\ref{thm:adap:regret} has an additional dependence on $\sqrt{\log (s-r)}$, it replaces the interval length $s-r$ with the cumulative loss over that interval, i.e., $\sum_{t=r}^{s}  f_t(\w)$. As a result, our bound could be much tighter when the comparator has a small loss. Whether the additional $\sqrt{\log (s-r)}$ factor can be removed remains an open problem to us, and we leave it as a future work.

\begin{figure*}
\centering
\begin{tabular}{@{}c@{\hspace{0.2ex}}*{23}{@{\hspace{0.1ex}}c}@{\hspace{0.2ex}}c@{}}
$t$ & 1 & 2 & 3 & 4 & 5 & 6 & 7 & 8 & 9 & 10 &11 &12 & 13 & 14 &15 & 16 & 17 & 18 & 19 & 20 & 21 & 22 & 23& $\cdots$ \\
 & $s_1$ &  & $s_2$ & $s_3$ &  &  & $s_4$ & $s_5$ &   & $s_6$ &$s_7$ &$s_8$ &  & $s_9$ &$s_{10}$ &  & $s_{11}$  & $s_{12}$  & $s_{13}$  & & $s_{14}$  & $s_{15}$  & $s_{16}$  & $\cdots$ \\
$\widetilde{\I}_0$ & [\quad \phantom{]} & \phantom{[}\quad ] &  [\quad ] & [\quad \phantom{]} & \phantom{[}\quad \phantom{]} & \phantom{[}\quad ] & [\quad ] & [\quad \phantom{]} & \phantom{[}\quad ] &  [\quad ] &[\quad ] &[\quad \phantom{]} & \phantom{[}\quad ] & [\quad ] & [\quad \phantom{]}& \phantom{[}\quad ]& [\quad ] &[\quad ] & [\quad \phantom{]} & \phantom{[}\quad ]  & [\quad ] &  [\quad ] & [\quad \phantom{]}&$\cdots$   \\
 $\widetilde{\I}_1$  &  &  &  [\quad \phantom{]} & \phantom{[}\quad \phantom{]} & \phantom{[}\quad \phantom{]} & \phantom{[}\quad ] & [\quad \phantom{]} & \phantom{[}\quad \phantom{]} & \phantom{[}\quad ] &  [\quad \phantom{]} &\phantom{[}\quad ] &[\quad \phantom{]} & \phantom{[}\quad \phantom{]} & \phantom{[}\quad ] & [\quad \phantom{]}& \phantom{[}\quad \phantom{]}& \phantom{[}\quad ] &[\quad \phantom{]} & \phantom{[}\quad \phantom{]} & \phantom{[}\quad ]  & [\quad \phantom{]} &  \phantom{[}\quad ]& [\quad \phantom{]} & $\cdots$   \\
  $\widetilde{\I}_2$ &  &  &      &     &     &  & [\quad \phantom{]} & \phantom{[}\quad \phantom{]} & \phantom{[}\quad \phantom{]} &  \phantom{[}\quad \phantom{]} &\phantom{[}\quad ] &[\quad \phantom{]} & \phantom{[}\quad \phantom{]} & \phantom{[}\quad \phantom{]} & \phantom{[}\quad \phantom{]}& \phantom{[}\quad \phantom{]}& \phantom{[}\quad ] &[\quad \phantom{]} & \phantom{[}\quad \phantom{]} & \phantom{[}\quad \phantom{]}  & \phantom{[}\quad \phantom{]} &  \phantom{[}\quad ] & [\quad \phantom{]} &$\cdots$   \\
 $\widetilde{\I}_3$ &  &  &      &     &     &  & &  &  &   &&[\quad \phantom{]} & \phantom{[}\quad \phantom{]} & \phantom{[}\quad \phantom{]} & \phantom{[}\quad \phantom{]}& \phantom{[}\quad \phantom{]}& \phantom{[}\quad \phantom{]} &\phantom{[}\quad \phantom{]} & \phantom{[}\quad \phantom{]} & \phantom{[}\quad \phantom{]}  & \phantom{[}\quad \phantom{]} &  \phantom{[}\quad ] & [\quad \phantom{]}&$\cdots$   \\
$\widetilde{\I}_4$ &  &  &      &     &     &  & &  &  &   && &  & & & &  & &  &   &  & & [\quad \phantom{]}&$\cdots$
\end{tabular}\vspace{-2ex}
\caption{Problem-dependent geometric covering (PGC) intervals. In the figure, each interval is denoted by $[ \quad ]$.}
\label{fig:interval:PGC}  \vspace{2ex}
\centering
\begin{tabular}{@{}c@{\hspace{0.2ex}}*{23}{@{\hspace{0.1ex}}c}@{\hspace{0.2ex}}c@{}}
$t$ & 1 & 2 & 3 & 4 & 5 & 6 & 7 & 8 & 9 & 10 &11 &12 & 13 & 14 &15 & 16 & 17 & 18 & 19 & 20 & 21 & 22 & 23& $\cdots$ \\
 & $s_1$ &  & $s_2$ & $s_3$ &  &  & $s_4$ & $s_5$ &   & $s_6$ &$s_7$ &$s_8$ &  & $s_9$ &$s_{10}$ &  & $s_{11}$  & $s_{12}$  & $s_{13}$  & & $s_{14}$  & $s_{15}$  & $s_{16}$  & $\cdots$ \\
$\widetilde{\C}_0$ & [\quad \phantom{]} & \phantom{[}\quad ] &  & [\quad \phantom{]} & \phantom{[}\quad \phantom{]} & \phantom{[}\quad ] &  & [\quad \phantom{]} & \phantom{[}\quad ] &   &[\quad ] & &  & [\quad ] & & & [\quad ] & & [\quad \phantom{]} & \phantom{[}\quad ]  &  &  [\quad ] & &$\cdots$   \\
 $\widetilde{\C}_1$  &  &  &  [\quad \phantom{]} & \phantom{[}\quad \phantom{]} & \phantom{[}\quad \phantom{]} & \phantom{[}\quad ] &  &  &  &  [\quad \phantom{]} &\phantom{[}\quad ] &&  & & [\quad \phantom{]}& \phantom{[}\quad \phantom{]}& \phantom{[}\quad ] &&  &   & [\quad \phantom{]} &  \phantom{[}\quad ]& & $\cdots$   \\
  $\widetilde{\C}_2$ &  &  &   &     &     &  & [\quad \phantom{]} & \phantom{[}\quad \phantom{]} & \phantom{[}\quad \phantom{]} &  \phantom{[}\quad \phantom{]} &\phantom{[}\quad ] & &  & & & &  &[\quad \phantom{]} & \phantom{[}\quad \phantom{]} & \phantom{[}\quad \phantom{]}  & \phantom{[}\quad \phantom{]} &  \phantom{[}\quad ] &  &$\cdots$   \\
 $\widetilde{\C}_3$ &  &  &      &     &     &  & &  &  &   &&[\quad \phantom{]} & \phantom{[}\quad \phantom{]} & \phantom{[}\quad \phantom{]} & \phantom{[}\quad \phantom{]}& \phantom{[}\quad \phantom{]}& \phantom{[}\quad \phantom{]} &\phantom{[}\quad \phantom{]} & \phantom{[}\quad \phantom{]} & \phantom{[}\quad \phantom{]}  & \phantom{[}\quad \phantom{]} &  \phantom{[}\quad ] & &$\cdots$   \\
  $\widetilde{\C}_4$ &  &  &      &     &     &  & &  &  &   && &  & & & &  & &  &   &  & & [\quad \phantom{]}&$\cdots$
\end{tabular}\vspace{-2ex}
\caption{Compact problem-dependent geometric covering (CPGC) intervals. In the figure, each interval is denoted by $[ \quad ]$.}
\label{fig:interval:CPGC}
\end{figure*}
\subsection{Problem-dependent Intervals}
We can refer to our result in Theorem~\ref{thm:adap:regret} as a \emph{problem-dependent} bound, since the dominant factor $\sqrt{\sum_{t=r}^{s}  f_t(\w)}$ depends on the problem, which has a similar spirit with the \emph{data-dependent} bound of Adagrad \citep{JMLR:Adaptive}. One unsatisfactory point of Theorem~\ref{thm:adap:regret} is that the logarithmic factor $\log s  \cdot \log (s-r)$, although non-dominant, is problem-independent. In this section, we discuss how to make SACS fully problem-dependent.

The problem-independent factor appears because CGC intervals, as well as CG intervals, are problem-independent. To address this limitation, we propose a problem-dependent way to generate intervals dynamically. The basic idea is to run an instance of SOGD, and restart the algorithm when the cumulative loss is larger than some threshold. The time points when SOGD restarts will be used as the starting rounds of intervals.

Specifically, we set $s_1=1$ and run an instance of SOGD. Let $s_1+\alpha$ be the round such that the cumulative loss  becomes larger than a threshold $C$. Then, we set $s_2=s_1+ \alpha+1$ and restart SOGD in round $s_2$. Repeating this process, we can generate a sequence of points $s_1,s_2,s_3,\ldots$ which is  referred to as  \emph{markers}.  Our problem-dependent geometric covering (PGC) intervals are constructed based on  markers:
\[
 \widetilde{\I}= \bigcup_{k \in \N \cup \{0\}}  \widetilde{\I}_k
 \]
 where for all $k \in \N \cup \{0\}$
 \[
 \widetilde{\I}_k=\left\{ [s_{i \cdot 2^k}, s_{(i+1) \cdot 2^k} -1]: i \in \N\right\}.
 \]
Similarly, we can also compact PGC intervals by removing overlapping intervals with the same starting point. The compact problem-dependent geometric covering (CPGC) intervals are given by
\begin{equation} \label{eqn:CPGC}
 \widetilde{\C}= \bigcup_{k \in \N \cup \{0\}}  \widetilde{\C}_k
\end{equation}
 where for all $k \in \N \cup \{0\}$
 \[
 \widetilde{\C}_k=\left\{ [ s_{i \cdot 2^k}, s_{(i+1) \cdot 2^k} -1]: i \textrm{ is odd}\right\}.
 \]
We provide graphical illustrations of PGC intervals and CPGC intervals in Fig.~\ref{fig:interval:PGC} and Fig.~\ref{fig:interval:CPGC}, respectively.

To see the difference between problem-independent and problem-dependent intervals, let's compare Fig.~\ref{fig:interval:saol} of GC intervals and Fig.~\ref{fig:interval:PGC} of PGC intervals. We have the following observations.
 \begin{compactitem}
 \item In the former one, intervals belong to the same level, i.e., $\I_k$, are of the same length, while in the latter one, intervals belong to the same level, i.e., $\widetilde{\I}_k$, are of different lengths.
 \item In the former one, an interval is created for each round. By contrast, in the latter one, an interval is created only at markers. Thus, the number of problem-dependent intervals is smaller than that of problem-independent intervals.
\end{compactitem}
We then incorporate CPGC intervals into our SACS algorithm, and summarize the procedure in Algorithm~\ref{alg:SACS:CPGC}. The new algorithm is a bit more complex than the original one in Algorithm~\ref{alg:SACS} because we need to construct CPGC intervals on the fly.

\begin{algorithm}[tb]
   \caption{SACS with CPGC intervals}
   \label{alg:SACS:CPGC}
\begin{algorithmic}[1]
   \STATE {\bfseries Input:} Parameter $C$
   \STATE Initialize indicator $NewInterval = true$, the total number of intervals $m=0$, the index of the latest interval $n=0$
   \FOR{$t=1$ {\bfseries to} $T$}
   \IF{$NewInterval$ is $true$}
   \STATE Initialize an expert $E_t$ by invoking SOGD in Algorithm~\ref{alg:SOGD}  and set $R_{t-1,t} =  C_{t-1,t}=0$
   \STATE Add $E_t$ to the set of active experts
   \[
      \S_t=\S_{t-1} \cup \{E_t \}
   \]
   \STATE Reset the indicator $NewInterval = false$
   \STATE Update the total number of intervals $m=m+1$
   \STATE Set $g_t=j$ such that $[m,j-1] \in \C$
    \STATE Record the index of the latest expert $n=t$
   \STATE Initialize the cumulative loss $L_{t-1}=0$
   \ENDIF
   \STATE Receive the prediction $\w_{t,i}$ of each expert $E_i \in \S_t$, and calculate its weight $p_{t,i}$ according to (\ref{eqn:weight})
    \STATE Submit $\w_t$ defined in (\ref{eqn:wt}) and then receive $f_t(\cdot)$
   \STATE   Update the cumulative loss of the latest expert $E_n$
   \[
   L_t=L_{t-1}+ f_t(\w_{t,n})
   \]
   \IF{$L_t > C$}
    \STATE Set the indicator $NewInterval = true$
    \STATE Remove experts whose ending times are $t+1$
    \[
    \S_{t} = \S_{t} \setminus \{ E_i | g_i = m+1 \}
    \]
   \ENDIF
   \STATE For each $E_i \in \S_t$, update
   \[
   \begin{split}
   R_{t,i} = R_{t-1,i} + f_t(\w_t) - f_t(\w_{t,i}), \textrm{ and }     C_{t,i} = C_{t-1,i} + \left|f_t(\w_t) - f_t(\w_{t,i})\right|
   \end{split}
   \]
   \STATE Pass $f_t(\cdot)$ to each expert $E_i \in \S_t$
   \ENDFOR
\end{algorithmic}
\end{algorithm}

Next, we explain the main differences. To generate CPGC intervals dynamically, we introduce a Boolean variable $NewInterval$ to indicate whether a new interval should be created, $m$ to denote the total number of intervals created so far, and $n$ to denote the index of the latest interval. In each round $t$, if $NewInterval$  is true, we will create a new expert $E_t$, add it to the active set, and then reset the indicator (Steps 5 to 7). We also increase the total number of intervals by $1$ in Step 8, and note that the $m$-th marker $s_m=t$. Let $m=i \cdot 2^k$, where $i$ is odd and $k\in \N$. According to the definition of CPGC intervals, $E_t=E_{s_m}$ is active during the interval $[ s_{i \cdot 2^k}, s_{(i+1) \cdot 2^k} -1]$. So, it should be removed before the $s_{(i+1) \cdot 2^k}$-th round. However, the value of $s_{(i+1) \cdot 2^k}$ is \emph{unknown} in the $t$-th round, so we cannot tag the ending time to $E_t$. As an alternative, we record the value of $(i+1) \cdot 2^k$, denoted by $g_t$ (Step 9), and remove $E_t$ when $m$ is going to reach $g_t$ (Step 18).

To generate the next marker $s_{m+1}$, we keep track of the index of the latest expert (Step 10), and record its cumulative loss (Steps 11 and 15). When the cumulative loss is larger than the threshold $C$ (Step 16), we set the indicator $NewInterval$ to be true (Step 17) and remove all the experts whose ending times are $s_{m+1}-1$ (Step 18). All the other steps are identical to those in Algorithm~\ref{alg:SACS}.

We present theoretical guarantees of  Algorithm~\ref{alg:SACS:CPGC}. As before, we first prove the meta-regret.
\begin{lemma}\label{lem:meta:CPGC} Suppose
\begin{equation} \label{eqn:C:bound}
C  \geq 20 H D^2 + 2D\sqrt{2 \delta} .
\end{equation}
Under Assumptions \ref{ass:4} and \ref{ass:6}, for any interval $[i, j] \in \widetilde{\C}$, and any $t \in [i,j]$, SACS with CPGC intervals satisfies
\[
\sum_{u=i}^{t}  \left[f_u(\w_u)-f_u(\w_{u,i})\right] \leq \tilde{c}(t) + \sqrt{2 \tilde{c}(t)\sum_{u=i}^{t} f_u(\w_{u,i})}
\]
where
\begin{equation} \label{eqn:tilde:c}
\begin{split}
 \tilde{c}(t)\leq  3 \ln  \left( 1 + \frac{4}{C}\sum_{u=1}^{t}    f_u(\w) \right) + 3\ln \frac{5+3\ln(1+t)}{2} .
\end{split}
\end{equation}
\end{lemma}
\textbf{Remark:} Following previous studies \citep{Chernov:2010:PAU,pmlr-v40-Luo15}, we treat the double logarithmic factor in $\tilde{c}(t)$ as a constant. Compared with Lemma~\ref{lem:meta}, the main advantage is that $c(t)$ is replaced with a problem-dependent term $\tilde{c}(t)$.

Based on Lemma~\ref{lem:meta:CPGC} and Theorem~\ref{thm:smooth:2}, we  prove a counterpart of Lemma~\ref{lem:regret:special}, which bounds the regret over any interval in $\widetilde{\C}$.
\begin{lemma}\label{lem:regret:special:CPGC} Under condition (\ref{eqn:C:bound}) and  Assumptions~\ref{ass:3}, \ref{ass:4}, \ref{ass:5} and \ref{ass:6}, for any interval $[i, j] \in \widetilde{\C}$,  any $t \in [i,j]$, and any $\w \in \W$, SACS with CPGC intervals satisfies
\[
\sum_{u=i}^{t}  \left[f_u(\w_u)-    f_u(\w) \right]\leq \tilde{a}(t)   +\sqrt{\tilde{b}(t) \sum_{u=i}^t    f_u(\w)}
\]
where
\begin{align}
\tilde{a}(t)= & \frac{3}{2} \tilde{c}(t) + 18 H D^2  + 2D\sqrt{2 \delta}, \label{eqn:at:tilde}\\
\tilde{b}(t)= &8\tilde{c}(t) +16 H D^2, \label{eqn:bt:tilde}
\end{align}
and $\tilde{c}(t)$ conforms to (\ref{eqn:tilde:c}).
\end{lemma}
Finally, we extend Lemma~\ref{lem:regret:special:CPGC} to any interval $[r,s] \subseteq [T]$.
\begin{thm} \label{thm:adap:regret:CPGC} Under condition (\ref{eqn:C:bound}) and Assumptions~\ref{ass:3}, \ref{ass:4}, \ref{ass:5} and \ref{ass:6}, for any interval $[r,s] \subseteq [T]$ and any $\w \in \W$, SACS with CPGC intervals satisfies
\[
\begin{split}
\sum_{t=r}^{s} \left[f_t(\w_t) -   f_t(\w)\right] \leq& 2(C+1) +\frac{3}{2} \tilde{c}(s) +  v \tilde{a}(s) + \sqrt{v \tilde{b}(s) \sum_{t=r}^{s}  f_t(\w)}\\
= & O\left(  \sqrt{\left(\sum_{t=r}^{s}  f_t(\w) \right)\log  \sum_{t=1}^{s}  f_t(\w)  \cdot \log \sum_{t=r}^{s}    f_t(\w) }  \right)
\end{split}
\]
where
\[
v \leq \left\lceil \log_2 \left(2 +\frac{4}{C} \sum_{t=r}^{s}    f_t(\w) \right) \right\rceil
\]
 $\tilde{a}(\cdot)$,  $\tilde{b}(\cdot)$ and $\tilde{c}(\cdot)$ are respectively defined in (\ref{eqn:at:tilde}), (\ref{eqn:bt:tilde}), and (\ref{eqn:tilde:c}).
\end{thm}
\textbf{Remark:} Compared with the upper bound in Theorem~\ref{thm:adap:regret}, we observe that the problem-independent term $\log s  \cdot \log (s-r)$ is improved to $\log  \sum_{t=1}^{s}  f_t(\w)  \cdot \log \sum_{t=r}^{s}    f_t(\w)$. As a result, our SACS with CPGC intervals becomes fully problem-dependent.

\section{Analysis}
In this section, we provide proofs of our key lemmas and theorems. The omitted ones are provided in the appendices.
\subsection{Proof of Theorem~\ref{thm:smooth:2}}
\citet{Scale:Free:Online} have analyzed the regret bound of  SOGD for online linear optimization. For the sake of completeness, we first present the proof of their regret bound, and then refine it by exploiting smoothness.

Define $\w_{t+1}'=\w_t - \eta_t \nabla f_t(\w_t)$. For any $\w \in \W$, we have
\[
\begin{split}
f_t(\w_{t}) - f_t(\w) \leq & \langle \nabla f_t(\w_{t}), \w_{t} - \w\rangle = \frac{1}{\eta_t} \langle \w_{t}  - \w_{t+1}', \w_{t} - \w\rangle \\
= & \frac{1}{2 \eta_t} \left( \|\w_t-\w\|^2 - \|\w_{t+1}'-\w\|^2 + \|\w_t  - \w_{t+1}'\|^2 \right) \\
=& \frac{1}{2 \eta_t} \left( \|\w_{t}-\w\|^2 - \|\w_{t+1}'-\w\|^2 \right) + \frac{\eta_t}{2 } \|\nabla f_t(\w_{t})\|^2.
\end{split}
\]
Summing the above inequality over all iterations, we have
\begin{equation}\label{eqn:SOGD:1}
\begin{split}
&\sum_{t=1}^T f_t(\w_{t}) - \sum_{t=1}^T  f_t(\w) \\
\leq & \frac{1}{2 \eta_1} \|\w_{1}-\w\|_2^2 + \sum_{t= 2}^T \left( \frac{1}{ \eta_t} - \frac{1}{ \eta_{t-1}} \right) \frac{\|\w_{t}-\w\|^2 }{2}+  \frac{1}{2} \sum_{t=1}^T\eta_t\|\nabla f_t(\w_{t})\|^2\\
\overset{\text{(\ref{eqn:domain})}}{\leq} &\frac{D^2}{2 \eta_1}   + \sum_{t= 2}^T \left( \frac{1}{ \eta_t} - \frac{1}{ \eta_{t-1}} \right) \frac{D^2}{2} + \frac{1}{2} \sum_{t=1}^T\eta_t\|\nabla f_t(\w_{t})\|^2=  \frac{D^2}{2 \eta_T} + \frac{1}{2} \sum_{t=1}^T\eta_t\|\nabla f_t(\w_{t})\|^2.
\end{split}
\end{equation}

To bound the last term of (\ref{eqn:SOGD:1}), we make use of the following lemma.
\begin{lemma}[Lemma 3.5 of \citet{AUER200248}] \label{lem:sum} Let $l_1,\ldots$, $l_T$ and $\delta$ be non-negative real numbers. Then
\[
\sum_{t=1}^T \frac{l_t}{\sqrt{\delta+\sum_{i=1}^t l_i}} \leq 2 \left( \sqrt{\delta + \sum_{t=1}^T l_t} -\sqrt{\delta}\right)
\]
where $0/\sqrt{0}=0$.
\end{lemma}

According to (\ref{eqn:eta_t}) and Lemma~\ref{lem:sum}, we have
\begin{equation}\label{eqn:SOGD:2}
\begin{split}
\sum_{t=1}^T \eta_t\|\nabla f_t(\w_{t})\|^2 =   \alpha \sum_{t=1}^T   \frac{ \|\nabla f_t(\w_{t})\|^2}{\sqrt{\delta+ \sum_{i=1}^t \|\nabla f_i(\w_{i})\|^2} }
\leq 2 \alpha  \sqrt{\delta+ \sum_{t=1}^T \|\nabla f_t(\w_{t})\|^2}.
\end{split}
\end{equation}
Substituting (\ref{eqn:SOGD:2}) into (\ref{eqn:SOGD:1}), we have
\begin{equation}\label{eqn:SOGD:3}
\begin{split}
&\sum_{t=1}^T f_t(\w_{t}) - \sum_{t=1}^T  f_t(\w)\\
\leq &   \left( \frac{D^2}{2 \alpha} +\alpha \right) \sqrt{\delta+ \sum_{t=1}^T \|\nabla f_t(\w_{t})\|^2} = \sqrt{2D^2} \sqrt{\delta+ \sum_{t=1}^T \|\nabla f_t(\w_{t})\|^2}
\end{split}
\end{equation}
where we set $\alpha= D/\sqrt{2}$.

Next, we introduce the self-bounding property of smooth functions \citep[Lemma 3.1]{NIPS2010_Smooth}.
\begin{lemma} \label{lem:smooth} For an $H$-smooth and nonnegative function $f: \W \mapsto \R$,
\[
\| \nabla f(\w)\| \leq \sqrt{4 H f(\w)}, \ \forall \w \in \W.
\]
\end{lemma}
From the analysis of \citet[Lemma 2.1 and Lemma 3.1]{NIPS2010_Smooth}, it is easy to see that the function $f$  needs to be nonnegative outside $\W$. So, in Assumption~\ref{ass:4}, we require $f_t(\cdot)$ is nonnegative over the whole space.
Combining Lemma~\ref{lem:smooth}, Assumptions~\ref{ass:4} and \ref{ass:5}, we have
\begin{equation} \label{eqn:smooth:key}
\|\nabla f_t(\w)\|^2 \leq 4 H  f_t (\w), \ \forall \w \in \W.
\end{equation}

From (\ref{eqn:SOGD:3}) and (\ref{eqn:smooth:key}), we have
\[
\sum_{t=1}^T f_t(\w_{t}) - \sum_{t=1}^T  f_t(\w) \leq  \sqrt{2D^2} \sqrt{ \delta+ 4 H \sum_{t=1}^T  f_t(\w_{t}) }=\sqrt{8 H D^2}\sqrt{\frac{\delta}{4H}+ \sum_{t=1}^T f_t(\w_t)}.
\]
To simplify the above inequality, we use the following lemma.
\begin{lemma}[Lemma 19 of \citet{Shai:thesis}] \label{lem:tool} Let $x, b,c \in \R_+$. Then,
\[
x  -c \leq b \sqrt{x} \Rightarrow  x - c \leq b^2 + b \sqrt{c}.
\]
\end{lemma}
Since
\[
\begin{split}
\left(\frac{\delta}{4H}+\sum_{t=1}^T f_t(\w_{t})  \right)- \left( \frac{\delta}{4H}+ \sum_{t=1}^T  f_t(\w) \right)\leq  \sqrt{8 H D^2}\sqrt{\frac{\delta}{4H}+ \sum_{t=1}^T f_t(\w_t)},
\end{split}
\]
Lemma~\ref{lem:tool} implies
\[
\begin{split}
\left(\frac{\delta}{4H}+\sum_{t=1}^T f_t(\w_{t})  \right)- \left( \frac{\delta}{4H}+ \sum_{t=1}^T  f_t(\w) \right)\leq & 8 H D^2 +\sqrt{8 H D^2\left(\frac{\delta}{4H}+ \sum_{t=1}^T  f_t(\w)\right)}\\
=& 8 H D^2 +D\sqrt{2 \delta + 8 H \sum_{t=1}^T  f_t(\w)}
\end{split}
\]
which completes the proof.

\subsection{Proof of Lemma~\ref{lem:meta}}
The lemma is obtained by tailoring Theorems 1 and 2 of \citet{pmlr-v40-Luo15} to our problem, and we present the proof to make the paper self-contained.  To this end, we first introduce the following lemma.
\begin{lemma}[Lemma 5 of \citet{pmlr-v40-Luo15}] \label{lem:phi} For any $R \in \R$, $C \geq 0$ and $r \in [-1,1]$, we have
\[
\Phi(R+r,C+|r|) \leq \Phi(R,C) + w(R,C) r + \frac{3 |r|}{ 2(C+1)}.
\]
\end{lemma}

For any $E_i \in \S_t$, i.e., $t \in [i,j] \in \C$ for certain $j$, we define
\[
r_{t,i}= f_t(\w_t) - f_t(\w_{t,i}).
\]
Then
\[
R_{t-1,i}  = \sum_{u=i}^{t-1} r_{u,i},  \textrm{ and } C_{t-1,i} = \sum_{u=i}^{t-1} \left|r_{u,i} \right|.
\]
According to Lemma~\ref{lem:phi}, we have
\[
\Phi(R_{t,i},C_{t,i}) \leq \Phi(R_{t-1,i},C_{t-1,i}) + w(R_{t-1,i},C_{t-1,i}) r_{t,i} + \frac{3 |r_{t,i}|}{ 2(C_{t-1,i}+1)}.
\]
Summing the above inequality over $E_i \in \S_t$, we have
\begin{equation} \label{eqn:meta:1}
\begin{split}
&\sum_{E_i \in \S_t} \Phi(R_{t,i},C_{t,i}) \\
\leq &\sum_{E_i \in \S_t} \Phi(R_{t-1,i},C_{t-1,i})+ \sum_{E_i \in \S_t} w(R_{t-1,i},C_{t-1,i}) r_{t,i} + \sum_{E_i \in \S_t} \frac{3 |r_{t,i}|}{ 2(C_{t-1,i}+1)}.
\end{split}
\end{equation}
We proceed by noticing that
\begin{equation} \label{eqn:meta:2}
\begin{split}
&\sum_{E_i \in \S_t} w(R_{t-1,i},C_{t-1,i}) r_{t,i} \overset{\text{(\ref{eqn:weight})}}{=}  \left(\sum_{E_i \in \S_t} w(R_{t-1,i},C_{t-1,i}) \right) \sum_{E_i \in \S_t} p_{t,i} \left(f_t(\w_t) -  f_t(\w_{t,i})\right)\\
=& \left(\sum_{E_i \in \S_t} w(R_{t-1,i},C_{t-1,i}) \right) \left( f_t(\w_t) -  \sum_{E_i \in \S_t}  p_{t,i} f_t(\w_{t,i})\right) \leq 0
\end{split}
\end{equation}
where the last step is due to $\w_t=\sum_{E_i \in \S_t} p_{t,i}  \w_{t,i}$ and Jensen's inequality \citep{Convex-Optimization}.

Combining (\ref{eqn:meta:1}) and (\ref{eqn:meta:2}), we arrive at
\begin{equation} \label{eqn:meta:3}
\sum_{E_i \in \S_t} \Phi(R_{t,i},C_{t,i}) \leq \sum_{E_i \in \S_t} \Phi(R_{t-1,i},C_{t-1,i})+  \sum_{E_i \in \S_t} \frac{3 |r_{t,i}|}{ 2(C_{t-1,i}+1)}.
\end{equation}

Given an expert $E_i$, we denote its ending time by $e_i$, i.e.,
\[
e_i=\{j:[i, j] \in \C\}.
\]
Summing  (\ref{eqn:meta:3}) over iterations $1,\ldots, t$ and simplifying, we have
\begin{equation} \label{eqn:meta:4}
\begin{split}
&\sum_{i=1}^t  \Phi(R_{t \wedge e_i,i},C_{t \wedge e_i,i})\\
 \leq & \sum_{i=1}^t\Phi(R_{i-1,i}, C_{i-1,i}) + \frac{3}{2} \sum_{i=1}^t \sum_{j=i}^{t \wedge e_i} \frac{ |r_{j,i}|}{ (C_{j-1,i}+1)}\\
=&\sum_{i=1}^t\Phi(0, 0)+ \frac{3}{2} \sum_{i=1}^t \sum_{j=i}^{t \wedge e_i} \frac{ |r_{j,i}|}{ (C_{j-1,i}+1)} =t + \frac{3}{2} \sum_{i=1}^t \sum_{j=i}^{t \wedge e_i} \frac{ |r_{j,i}|}{ (C_{j-1,i}+1)}
\end{split}
\end{equation}
where $t \wedge e_i = \min(t, e_i)$. To bound the last term, we make use of the following lemma.
\begin{lemma}[Lemma 14 of \citet{pmlr-v35-gaillard14}] \label{lem:sum:inverse}
Let $a_0 > 0$ and $a_1, \ldots, a_m \in [0,1]$ be real numbers and let $f: (0,+\infty) \mapsto [0, +\infty)$
be a nonincreasing function. Then
\[
\sum_{i=1}^m a_i f(a_0+\cdots + a_{i-1}) \leq f(a_0) + \int_{a_0}^{a_0+\cdots + a_{m}} f(x) d x.
\]
\end{lemma}
Applying Lemma~\ref{lem:sum:inverse} with $f(x)=1/x$, we have
\begin{equation} \label{eqn:meta:5}
\sum_{j=i}^{t \wedge e_i} \frac{ |r_{j,i}|}{ (C_{j-1,i}+1)} \leq 1 +  \int_{1}^{1+C_{t \wedge e_i,i}} \frac{1}{x} d x = 1+ \ln(1+C_{t \wedge e_i,i}) .
\end{equation}

Substituting (\ref{eqn:meta:5}) into (\ref{eqn:meta:4}), we obtain
\[
\sum_{i=1}^t  \Phi(R_{t \wedge e_i,i},C_{t \wedge e_i,i}) \leq \frac{5}{2} t + \frac{3}{2}\sum_{i=1}^t \ln(1+C_{t \wedge e_i,i}) \leq  t \left( \frac{5}{2} + \frac{3}{2} \ln(1+t) \right).
 \]
Thus, for any $E_i \in \S_t$, we have
\[
\Phi(R_{t,i},C_{t,i})=\Phi(R_{t \wedge e_i,i},C_{t \wedge e_i,i}) \leq t \left( \frac{5}{2} + \frac{3}{2} \ln(1+t) \right) \leq 4 t^2
\]
implying
\begin{equation} \label{eqn:meta:6}
R_{t,i} \leq  \sqrt{  3 \ln(4t^2) C_{t,i} } .
\end{equation}

Define
\[
D_{t,i}= \sum_{u=i}^{t} \left[ f_u(\w_{u,i})  - f_u(\w_u)\right]_+ , \textrm{ and } L_{t,i}=\sum_{u=i}^{t} f_u(\w_{u,i}).
\]
It is easy to verify that
\[
C_{t,i}=  \sum_{u=i}^{t} \left|r_{u,i} \right|=\sum_{u=i}^{t} \left(r_{u,i} +  2 \left[ -r_{u,i}\right]_+\right) = R_{t,i} +2  D_{t,i} \leq R_{t,i} +2  L_{t,i}.
\]
Plugging the above inequality into (\ref{eqn:meta:6}), we have
\[
R_{t,i} \leq  \sqrt{ 3 \ln(4t^2) \left( R_{t,i} +2  L_{t,i}\right) } .
\]

Consider the case $R_{t,i} \geq 0$. Then, the above inequality implies
\[
R_{t,i}^2 - 3 \ln(4t^2)  R_{t,i}  -6 \ln(4t^2)   L_{t,i} \leq 0,
\]
thus
\[
R_{t,i} \leq \frac{3 \ln(4t^2)  + \sqrt{[3 \ln(4t^2)]^2 + 24  \ln(4t^2) L_{t,i} }}{2} \leq 3 \ln(4t^2) + \sqrt{6 \ln(4t^2)   L_{t,i} }.
\]
We complete the proof by noticing the above inequality also holds for $R_{t,i} \leq 0$.
\subsection{Proof of Lemma~\ref{lem:regret:special}}
Notice that $E_i$ is an instance of SOGD that starts to work from round $i$. According to Theorem~\ref{thm:smooth:2}, we have
\begin{equation} \label{eqn:regret:special:1}
\sum_{u=i}^t f_u(\w_{u,i}) -  \sum_{u=i}^t    f_u(\w)  \leq   8 H D^2 +D\sqrt{2 \delta + 8 H \sum_{u=i}^t    f_u(\w)}
\end{equation}
for any $\w \in \W$. From (\ref{eqn:regret:special:1}), we can prove
\begin{equation} \label{eqn:regret:special:2}
\sum_{u=i}^t f_u(\w_{u,i}) -  \sum_{u=i}^t    f_u(\w)  \leq   10 H D^2 + D\sqrt{2 \delta} + \sum_{u=i}^t    f_u(\w).
\end{equation}

Combining the (\ref{eqn:regret:special:1}) with  Lemma~\ref{lem:meta}, we have
\[
\begin{split}
&\sum_{u=i}^{t}  f_u(\w_u)-\sum_{u=i}^t    f_u(\w) \\
\leq & c(t) + \sqrt{2 c(t)\sum_{u=i}^{t} f_u(\w_{u,i})}+8 H D^2 +D\sqrt{2 \delta + 8 H \sum_{u=i}^t    f_u(\w)}\\
\overset{\text{(\ref{eqn:regret:special:2})}}{\leq}  & c(t) + \sqrt{2 c(t)\left(10 H D^2 + D\sqrt{2 \delta}+ 2\sum_{u=i}^t    f_u(\w) \right)}+8 H D^2 +D\sqrt{2 \delta + 8 H \sum_{u=i}^t    f_u(\w)} \\
\leq & c(t)  + \sqrt{2 c(t) \left(10 H D^2 + D\sqrt{2 \delta} \right) }+ 8 H D^2 + D\sqrt{2 \delta} + \left(\sqrt{4 c(t)} + \sqrt{8 H D^2} \right) \sqrt{ \sum_{u=i}^t    f_u(\w)} \\
\leq & \frac{3}{2}c(t) + 18 H D^2  + 2D\sqrt{2 \delta}  +  \sqrt{\big(8 c(t)+16 HD^2 \big) \sum_{u=i}^t    f_u(\w)}.
\end{split}
\]

\subsection{Proof of Theorem~\ref{thm:adap:regret}}
For any interval $[r,s] \subseteq [T]$, it can be covered by a small number of intervals in $\C$. Specifically, we have the following property of CGC intervals, which is similar to Lemma 1.2 of \citet{Adaptive:ICML:15} and Lemma 1 of \citet{Dynamic:Regret:Adaptive}.

\begin{lemma}\label{lem:seqence} Let $[r,s] \subseteq [T]$ be an arbitrary interval. Then, we can find a sequence of consecutive intervals
 \[
 I_1=[i_1,i_2-1], \  I_2=[i_2,i_3-1], \ \ldots ,  \ I_v=[i_v,i_{v+1}-1] \in \C
 \]
 such that
 \[
 i_1=r,  \  i_v \leq s \leq i_{v+1}-1, \textrm{ and } v \leq \lceil \log_2 (s-r+2)\rceil .
 \]
\end{lemma}

Then, for the first $v-1$ intervals, Lemma~\ref{lem:regret:special} implies
\[
\begin{split}
\sum_{t=i_k}^{i_{k+1}-1}  f_t(\w_t)-\sum_{t=i_k}^{i_{k+1}-1}  f_t(\w) \leq & a(i_{k+1}-1)   +\sqrt{b(i_{k+1}-1) \sum_{t=i_k}^{i_{k+1}-1}    f_t(\w)} \\
\leq & a(s)   +\sqrt{b(s)\sum_{t=i_k}^{i_{k+1}-1}    f_t(\w)} ,  \ \forall k \in [v-1].
\end{split}
\]
And for the last interval, we have
\[
\sum_{t=i_v}^{s}  f_t(\w_t)-\sum_{t=i_v}^{s}    f_t(\w) \leq a(s)   +\sqrt{b(s) \sum_{t=i_v}^{s}    f_t(\w)}.
\]
By adding them together, we have
\[
\begin{split}
\sum_{t=r}^{s}  f_t(\w_t)-\sum_{t=r}^{s}  f_t(\w) \leq & v a(s)   + \sqrt{b(s)} \left(\sum_{k=1}^{v-1} \sqrt{\sum_{t=i_k}^{i_{k+1}-1}    f_t(\w)} +\sqrt{ \sum_{t=i_v}^{s}    f_t(\w)}\right)\\
\leq & v a(s) + \sqrt{v b(s) \sum_{t=r}^{s}  f_t(\w)}
\end{split}
\]
where the last step is due to the Cauchy--Schwarz inequality.
\subsection{Proof of Lemma~\ref{lem:meta:CPGC}}
The analysis is similar to the proof of Lemma~\ref{lem:meta}, with modifications that take into account the CPGC intervals. The key difference is that the total number of experts till round $t$ could be smaller than $t$, and is determined by the problem.

We first give an upper bound of the total number of experts created so far. Note that in each interval  $[s_i,s_{i+1}-1]$, an expert $E_{s_i}$ is created by running SOGD. According to Theorem~\ref{thm:smooth:2}, we have
\[
\begin{split}
\sum_{u=s_i}^{s_{i+1}-1} f_u(\w_{u,s_i}) -  \sum_{u=s_i}^{s_{i+1}-1}    f_u(\w)  \leq  & 8 H D^2 +D\sqrt{2 \delta + 8 H \sum_{u=s_i}^{s_{i+1}-1}    f_u(\w)}\\
\leq & 10 H D^2 + D\sqrt{2 \delta} + \sum_{u=s_i}^{s_{i+1}-1}    f_u(\w)
\end{split}
\]
for any $\w \in \W$. On the other hand, from the construction rule of markers, we have
\[
\sum_{u=s_i}^{s_{i+1}-1} f_u(\w_{u,s_i}) \geq C.
\]
Thus, we have
\begin{equation} \label{eqn:lower:loss}
\sum_{u=s_i}^{s_{i+1}-1}    f_u(\w) \geq \frac{1}{2} \big( C - (10 H D^2 + D\sqrt{2 \delta}) \big) \overset{\text{(\ref{eqn:C:bound})}}{\geq} \frac{C}{4}.
\end{equation}
Let $m$ be the number of experts created till round $t$. Summing (\ref{eqn:lower:loss}) over $i=1,\ldots,m-1$, we have
\[
\sum_{u=s_1}^{s_{m}-1}    f_u(\w) \geq \frac{C}{4} (m-1)
\]
implying
\begin{equation} \label{eqn:upper:m}
m \leq 1+ \frac{4}{C}\sum_{u=s_1}^{s_{m}-1}    f_u(\w) \leq 1 + \frac{4}{C}\sum_{u=1}^{t}    f_u(\w).
\end{equation}

Next, by repeating the analysis of Lemma~\ref{lem:meta}, we obtain (\ref{eqn:meta:3}). Then, we sum (\ref{eqn:meta:3}) over  iterations $1,\ldots, t$ and simplify to get
\begin{equation} \label{eqn:meta:CPGC:1}
\sum_{i=1}^m  \Phi(R_{t \wedge e_{s_i},s_i},C_{t \wedge e_{s_i},s_i}) \leq m + \frac{3}{2} \sum_{i=1}^m \sum_{j=s_i}^{t \wedge e_{s_i}} \frac{ |r_{j,s_i}|}{ (C_{j-1,s_i}+1)}
\end{equation}
where $t \wedge e_{s_i} = \min(t, e_{s_i})$, and $e_{s_i}$ is the ending time of expert $E_{s_i}$, i.e.,
\[
e_{s_i}=\{j:[s_i, j] \in \widetilde{\C}\}.
\]
Following the derivation of (\ref{eqn:meta:5}), we have
\begin{equation} \label{eqn:meta:CPGC:2}
\sum_{j=s_i}^{t \wedge e_{s_i}} \frac{ |r_{j,s_i}|}{ (C_{j-1,s_i}+1)}\leq 1+ \ln(1+C_{t \wedge e_{s_i},{s_i}}).
\end{equation}

Substituting (\ref{eqn:meta:CPGC:2}) into (\ref{eqn:meta:CPGC:1}), we obtain
\[
\sum_{i=1}^m  \Phi(R_{t \wedge e_{s_i},s_i},C_{t \wedge e_{s_i},s_i}) \leq \frac{5}{2} m  +  \frac{3}{2}\sum_{i=1}^m \ln(1+C_{t \wedge e_{s_i},{s_i}}) \leq \underbrace{m \left( \frac{5}{2} + \frac{3}{2}\ln(1+t) \right)}_{:=U}.
\]
Then, according to the rest analysis of Lemma~\ref{lem:meta}, for any $E_i \in \S_t$, we can prove
\[
R_{t,i} \leq  3 \ln(U) + \sqrt{6 \ln(U)   L_{t,i}}.
\]
\subsection{Proof of Lemma~\ref{lem:regret:special:CPGC}}
The proof is identical to that of Lemma~\ref{lem:regret:special} by replacing $c(t)$ with $\tilde{c}(t)$.

\subsection{Proof of Theorem~\ref{thm:adap:regret:CPGC}}
Let $s_p$ be the smallest marker that is larger than $r$, and $s_q$ be the largest marker that is not larger than $s$. Then, we have
\[
s_{p-1} \leq r < s_p , \textrm{ and }  s_{q} \leq s < s_{q+1}.
\]

First, we  bound the regret over interval $[r, s_p-1]$. We have
\[
\begin{split}
& \sum_{t=r}^{s_p-1} f_t(\w_{t}) -  \sum_{t=r}^{s_p-1} f_t(\w)  \leq \sum_{t=r}^{s_p-1} f_t(\w_{t}) \leq \sum_{t=s_{p-1}}^{s_p-1} f_t(\w_{t}) \\
\leq & \sum_{t=s_{p-1}}^{s_p-1} f_t(\w_{t}) - \sum_{t=s_{p-1}}^{s_p-1} f_t(\w_{t,s_{p-1}}) + \sum_{t=s_{p-1}}^{s_p-1} f_t(\w_{t,s_{p-1}}) \\
\leq & \tilde{c}(s_p-1) + \sqrt{2 \tilde{c}(s_p-1) \sum_{t=s_{p-1}}^{s_p-1} f_t(\w_{t,s_{p-1}})} + \sum_{t=s_{p-1}}^{s_p-1} f_t(\w_{t,s_{p-1}}) \\
 \leq & \frac{3}{2} \tilde{c}(s_p-1) + 2\sum_{t=s_{p-1}}^{s_p-1} f_t(\w_{t,s_{p-1}})
\end{split}
\]
where the penultimate inequality is due to Lemma~\ref{lem:meta:CPGC}. According to the construction rule of markers and Assumption~\ref{ass:6}, we have
\[
\sum_{t=s_{p-1}}^{s_p-1} f_t(\w_{t,s_{p-1}}) \leq C+1.
\]
Thus
\begin{equation} \label{eqn:regret:CPGC:1}
\sum_{t=r}^{s_p-1} f_t(\w_{t}) -  \sum_{t=r}^{s_p-1} f_t(\w)  \leq \frac{3}{2} \tilde{c}(s_p-1) + 2(C+1) \leq  \frac{3}{2} \tilde{c}(s) + 2(C+1).
\end{equation}

Next, we bound the regret over interval $[s_p, s]$. To this end, we introduce the following lemma, which is similar to Lemma~\ref{lem:seqence}, but limited to intervals that start and end with markers.
\begin{lemma}\label{lem:seqence:CPGC} Let $[s_p,s_q]  \subseteq [T]$ be an interval that starts from an marker $s_p$ and ends at another marker $s_q$. Then, we can find a sequence of consecutive intervals
 \[
 I_1=[s_{i_1},s_{i_2}-1], \ I_2=[s_{i_2},s_{i_3}-1],\ \ldots , \  I_{v}=[s_{i_v},s_{i_{v+1}}-1] \in \widetilde{\C}
 \]
 such that
 \[
 i_1=p, \ i_v \leq q < i_{v+1}, \textrm{ and } v \leq \lceil \log_2 (q-p+2)\rceil .
 \]
\end{lemma}
Note that
\[
q < i_{v+1} \Rightarrow q+1 \leq i_{v+1} \Rightarrow s_{q+1} -1 \leq s_{i_{v+1}}-1 \Rightarrow s \leq s_{i_{v+1}}-1.
\]
Thus, the interval $[s_p, s]$ is also covered by the sequence of intervals in the above lemma. Then, we can repeat the analysis of Theorem~\ref{thm:adap:regret}, and prove that
\begin{equation} \label{eqn:regret:CPGC:2}
\sum_{t=s_p}^{s}  f_t(\w_t)-\sum_{t=s_p}^{s}  f_t(\w) \leq  v \tilde{a}(s) + \sqrt{v \tilde{b}(s) \sum_{t=s_p}^{s}  f_t(\w)}.
\end{equation}
Combining (\ref{eqn:regret:CPGC:1}) with (\ref{eqn:regret:CPGC:2}), we have
\[
\sum_{t=r}^{s}  f_t(\w_t)-\sum_{t=r}^{s}  f_t(\w) \leq  \frac{3}{2} \tilde{c}(s) + 2(C+1) + v \tilde{a}(s) + \sqrt{v \tilde{b}(s) \sum_{t=r}^{s}  f_t(\w)}.
\]

Finally, we provide an upper bound of $q-p$, which is similar to the upper bound of $m$ in (\ref{eqn:upper:m}). We sum (\ref{eqn:lower:loss}) over $i=p,\ldots,q-1$ and obtain
\[
\sum_{t=s_p}^{s_{q}-1}    f_t(\w) \geq \frac{C}{4} (q-p)
\]
which implies
\[
q-p \leq \frac{4}{C} \sum_{t=s_p}^{s_{q}-1}  f_t(\w) \leq  \frac{4}{C} \sum_{t=r}^{s}    f_t(\w).
\]
\section{Conclusion and Future Work}
In this paper,  we propose a Strongly Adaptive algorithm for Convex and Smooth functions (SACS), which combines the strength of online gradient descent (OGD), geometric covering (GC) intervals, and AdaNormalHedge.  Let $L_r^s(\w)$ be the cumulative loss of a comparator $\w$ over an interval $[r,s]$. Theoretical analysis shows that the regret of SACS over any $[r,s]$ with respect to any $\w$ is $O(\sqrt{L_r^s(\w) \log s  \cdot \log (s-r)})$, which could be much smaller than the $O(\sqrt{(s-r) \log s})$ regret \citep{Improved:Strongly:Adaptive} when $L_r^s(\w)$ is small. Furthermore, we propose to construct problem-dependent intervals, and improve the regret bound to $O(\sqrt{L_r^s(\w) \log L_1^s(\w)  \cdot \log L_r^s(\w)})$.

One future work is to extend our results to exp-concave functions. For this type of functions,  there exist  efficient algorithms that achieve $O(d \log^2 T)$ adaptive regret \citep{Adaptive:Hazan}, which is unfortunately problem-independent. Note that the static regret of exp-concave functions can be improved by  smoothness \citep{Beyond:Logarithmic}. Thus, we can improve the adaptive regret of exp-concave functions by using the regret bound of \citet{Beyond:Logarithmic} and incorporating our problem-dependent intervals into the FLH algorithm of \citet{Adaptive:Hazan}. We will provide a detailed investigation in the future.
\bibliography{E:/MyPaper/ref}

\appendix

\section{Proof of Lemma~\ref{lem:seqence}}
From the definition of $\C$ in (\ref{eqn:CGC}) and the graphical illustration in Fig.~\ref{fig:interval:compact}, it is easy to verify the following property holds: For any consecutive intervals $I=[i,j-1]$ and $J=[j,k]$ in $\C$, we have $|J|\geq 2 |I|$.
According to (\ref{eqn:CGC}), we must have  $i=\alpha 2^p$, where $\alpha$ is odd and $p \in \N$. Then, $j= \alpha 2^p+2^p= \frac{\alpha+1}{2^q} 2^{p+q}$ where $q\geq 1$ is the largest integer such that $\frac{\alpha+1}{2^q}$ is odd. The length of $I$ is $2^p$ and the length of $J$ is $2^{p+q}$, and thus $|J|\geq 2 |I|$.

Given an arbitrary interval  $[r,s]$, we set the first interval $I_1=[i_1=r,i_2-1]$ to be the interval in $\C$ that starts from $r$. If $s \geq i_2$, we set the second interval $I_2=[i_2, i_3-1]$ to be the interval in $\C$ that starts from $i_2$. We repeat this process until $s$ is not larger  than the ending time of last interval. Recall the length of $I_1$ is at least $1$, and the length of successive intervals at least doubles. Thus, the total number of intervals is at most
\[
\min\{k +1| 1+2+\ldots+2^k \geq s-r+1, k \in \N\} \leq \lceil \log_2 (s-r+2)\rceil.
\]

\section{Proof of Lemma~\ref{lem:seqence:CPGC}}
The proof is similar to that of Lemma~\ref{lem:seqence}. First, we prove that for any $3$ consecutive intervals
\[
[s_a, s_b-1], [s_b, s_c-1], [s_c,s_d-1] \in \widetilde{\C}
\]
we must have
\[
c-b \geq 2 (b-a).
\]
According to (\ref{eqn:CPGC}), we have $a=\alpha 2^p$, where $\alpha$ is odd and $p \in \N$. Then, we have $b=(\alpha+1) 2^p=\frac{\alpha+1}{2^q} 2^{p+q}$ where $q \geq 1$ is the largest integer such that $\frac{\alpha+1}{2^q}$ is odd. As a result $c=(\frac{\alpha+1}{2^q} +1) 2^{p+q}=(\alpha+1)2^p + 2^{p+q}$. Finally, we have
\[
c-b=2^{p+q} \geq 2^{p+1} \geq 2 (b-a).
\]

Given an arbitrary interval $[s_p,s_q]$, we choose the first interval $I_1=[s_{i_1}=s_{p},s_{i_2}-1]$ as the interval in $\widetilde{\C}$ that starts from $s_p$. If $q \geq i_2$, we choose the second interval $I_2=[s_{i_2}, s_{i_3}-1]$ as the interval in $\widetilde{\C}$ that starts from $s_{i_2}$. We repeat this process until finding an interval  $I_{v}=[s_{i_v},s_{i_{v+1}}-1]$ such that $q<i_{v+1}$. Recall the condition $( i_{v+1}-i_v) \geq 2 (i_v - i_{v-1}) \geq \cdots \geq  (i_2 - i_1)$, and the fact $i_2 - i_1 \geq 1$, we have
\[
i_{v+1}= i_1 + \sum_{j=1}^v (i_{j+1}-i_j) \geq p + \sum_{j=1}^v 2^{j-1}= p + 2^v-1.
\]
Thus, to ensure $i_{v+1} > q$, it is sufficient to require
\[
 p + 2^v-2 \geq  q.
\]
So, $v$ is at most $\lceil \log_2 (q-p+2)\rceil$.
\end{document}